\documentclass{cta-author}

\usepackage{diagbox}
\usepackage{booktabs}
\usepackage{multirow}
\usepackage{adjustbox}
\usepackage{graphics}
\usepackage{threeparttable}

{}
{}
{}

\begin{document}


\title{On the Generalisation Capabilities of Fisher Vector based Face Presentation Attack Detection}

\author{\au{L\'azaro J. Gonz\'alez-Soler$^{1\corr}$}, \au{Marta Gomez-Barrero$^{2}$}, \au{Christoph Busch$^{1}$}}

\address{\add{1}{Biometrics and Internet Security Research Group, 
		Hochschule Darmstadt, Germany}
\add{2}{Hochschule Ansbach,Germany}
\email{lazaro-janier.gonzalez-soler@h-da.de}}

\begin{abstract}
    In the last decades, the broad development experienced by biometric systems has unveiled several threats which may decrease their trustworthiness. Those are attack presentations which can be easily carried out by a non-authorised subject to gain access to the biometric system. In order to mitigate those security concerns, most face Presentation Attack Detection techniques have reported a good detection performance when they are evaluated on known Presentation Attack Instruments (PAI) and acquisition conditions, in contrast to more challenging scenarios where unknown attacks are included in the test set. For those more realistic scenarios, the existing algorithms face difficulties to detect unknown PAI species in many cases. In this work, we use a new feature space based on Fisher Vectors, computed from compact Binarised Statistical Image Features histograms, which allow discovering semantic feature subsets from known samples in order to enhance the detection of unknown attacks. This new representation, evaluated for challenging unknown attacks taken from freely available facial databases, shows promising results: a BPCER100 under 17\% together with an AUC over 98\% can be achieved in the presence of unknown attacks. In addition, by training a limited number of parameters, our method is able to achieve state-of-the-art deep learning-based approaches for cross-dataset scenarios.   
\end{abstract}

\maketitle

\section{Introduction}
\label{sec:intro}

Based on the lemma ``forget about PIN and passwords, you are your key'', the deployment of biometric systems has continuously increased over the last decades~\cite{Galbally-PAD-FaceSurvey-2014}. Among different biometric modalities, the face has become the second most largely deployed characteristic right after fingerprints in terms of market quota~\cite{Market-Report-2008}. In addition, it has been widely adopted in most official identification documents such as ICAO-compliant biometric passport~\cite{Kundra-ePassportStudy-2014} or national ID cards. 

In spite of its advantages, face-based biometric systems are also vulnerable to attack presentations: the broad development experienced by numerous social networks (e.g., LinkedIn, Facebook, or Youtube) allows a non-authorised subject to easily download and re-use a photo or video of a target victim, which shall be impersonated. This way, he or she can gain access to several applications such as the bank account or can unlock smartphones and circumvent border controls, in which face recognition systems are commonly deployed. In fact, malicious attackers can also create new challenging and sophisticated attacks such as 3D masks~\cite{Manjani-PAD-SiliconeDetection-TIFS-2017}, makeup~\cite{Rathgeb-MakeupAttackDetection-ICPR-2020}, or even virtual reality~\cite{Xu-PAD-VirtualReality-2016} to decrease the trustworthiness of facial biometric systems. Moreover, the recent advances on the creation of synthetic videos, or deep fakes, also pose a serious threat for IT-security in general~\cite{Tolosana-PAD-DeepFake-2020}.


In order to address those security threats, several face Presentation Attack Detection (PAD) methods have been proposed. By assuming that bona fide presentations (BP) should be inherently opposed to attack presentations (APs), many PAD approaches have employed handcrafted features such as Local Binary Patterns~(LBP)~\cite{Xiong-PAD-UnkPADrgbImages-BTAS-2018,Peng-PAD-LBPEnsembleLearning-2020}, 
Histogram of Oriented Gradients~(HOG)~\cite{Agarwal-PAD-Multispectral-2017}, and Local Phase Quantization~(LPQ)~\cite{Raghavendra-PAD-MS-LPQ-2018,Agarwal-PAD-Multispectral-2017} to perform a binary classification (i.e., BP vs. AP) using e.g., Support Vector Machines (SVM)~\cite{Maatta-PAD-MicroTexture-2011,Boulkenafet-PAD-ColorTextureAnalysis-ICIP-2015,Peng-PAD-GuidedScaleTexture-2018} or Linear Discriminant Analysis (LDA)~\cite{Galbally-PAD-IQA-2013,Erdogmus-PAD-3dMasks-TIFS-2014} classifiers. More recently, with the large development experienced by deep learning techniques, numerous PAD approaches have successfully applied Convolutional Neural Networks (CNNs) for binary classification on facial PAD~\cite{Atoum-PAD-depthCNN-IJCB-2017,Qu-PAD-ShallowCNN2019}. 

In general, both handcrafted- and deep learning-based methods have reported a high detection performance for identifying Presentation Attack Instruments (PAIs) when both the PAI species (i.e., attack type) and acquisition conditions are known a priori: the so-called known attack scenario. However, CNNs still have certain drawbacks:

\begin{figure}[!t]
	\centering
	\includegraphics[width=0.99\linewidth]{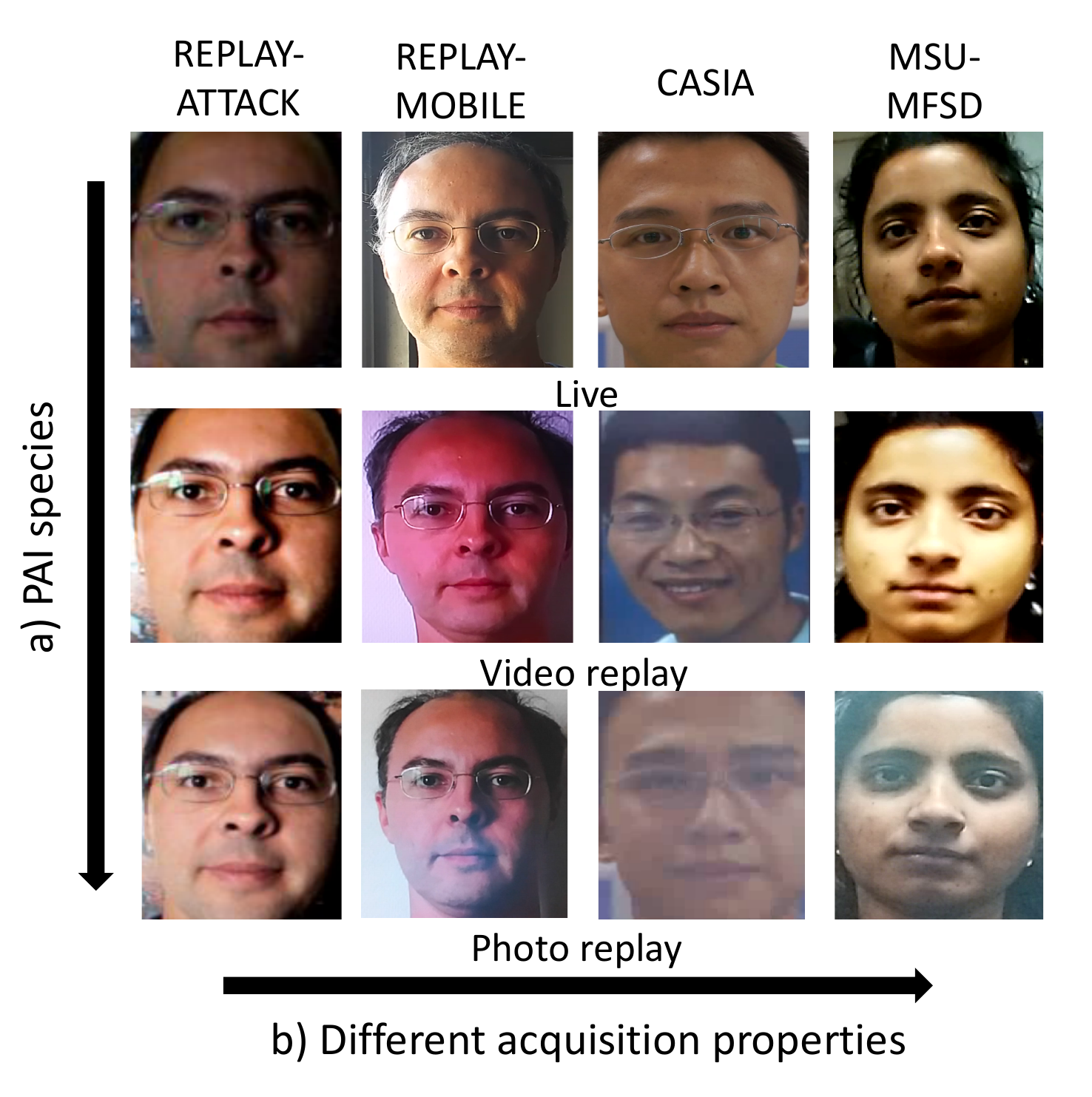}
	\caption{Two major generalisation shortcomings affecting current state-of-the-art PAD techniques: a) sophisticated attacks, which might be unknown for most PAD approaches and b) acquisition properties such as face appearance, pose, illumination, capture device, and subject vary from a dataset to another one.}
	\label{fig:attacks}
\end{figure}

\begin{itemize}

    \item \textit{A Large number of hyperparameters:} Deep learning-based approaches are usually based on dense CNNs with a large number of learnable parameters (exceeding 2.7 millions~\cite{Deb-PAD-GeneralizableFPAD-ArXiv-2020}).

    \item \textit{Poor generalisation capability across several datasets:} Most state-of-the-art PAD methods report an accuracy decrease when they are evaluated over a new database. Given that face capture devices might age and eventually stop working, the fabrication and re-capturing of an entire set of known PAI species with a new capture device might not be possible or at least require some time. Thus, the generalisation across several datasets is of utmost importance for the PAD testing task. Several studies have shown that differences between BP and AP samples across different datasets include aspects such skin detail, colour distortion, moir\'e pattern, shape deformation, and texture artefacts (see Fig.~\ref{fig:attacks}b), which widely vary among databases, hence leading to a poor PAD performance generalisation~\cite{Wang-PAD-AdvDomainAdaptation-TIFS-2020}. 

    \item \textit{Poor generalisation capability for unknown PAI species:} Current state-of-the-art techniques face difficulties to detect unknown PAI species (i.e., attack presentations created with a particular species type such as printed photo, 3D masks, and video replay, among others, which remain unknown for the detector during training), thereby resulting in a performance deterioration with respect to the detection of PAI species previously known in training.
\end{itemize}

In order to tackle those open issues, we focus on a different approach which has already shown remarkable results on challenging scenarios such as cross-dataset and unknown PAI species for fingerprint PAD~\cite{GonzalezSoler-FingerprintPADonLocalFeatures-IEEE-Access-2021}. In that work, a combination of local feature descriptors and global feature representation models a new feature space in which the generalisation capabilities of the PAD module are enhanced. In fact, that approach achieved the best detection accuracy in the LivDet 2019 competition~\cite{Orru-LivDet-ICB-2019}.  Whereas some keypoint based descriptors such as Scale-invariant feature transform (SIFT) and Speeded up robust features (SURF) have shown to be an appropriate choice for fingerprint samples~\cite{GonzalezSoler-FingerprintPADonLocalFeatures-IEEE-Access-2021,Gonzalez-PAD-MaterialImpact-ICB-2019}, in which minutiae can be regarded as landmarks within the image, for facial images the textural information is more relevant than the geometric details related to facial landmarks~\cite{Gonzalez-PAD-FVencForFacePAD-BIOSIG-2020}. Therefore, we use a new face PAD approach, which encodes accurate and compact dense Binarized Statistical Image Features (BSIF), extracted from local patches of the facial image, and projects them into a new feature space with Fisher Vectors (FV)~\cite{Gonzalez-PAD-FVencForFacePAD-BIOSIG-2020}. 

The main contributions of this work with respect to our preliminary study~\cite{Gonzalez-PAD-FVencForFacePAD-BIOSIG-2020} can be summarised as follows:

\begin{itemize}

    \item An extended study of the impact of different numbers of semantic sub-groups for facial PAD.    

    \item An analysis of the detection performance of our representation over three different colour spaces, namely RGB, HSV, and $\mathrm{YC_{b}C_{r}}$, in compliance with the ISO/IEC 30107-3 evaluation metrics for biometric PAD~\cite{ISO-IEC-30107-3-PAD-metrics-170227}.  

    \item  A thorough analysis on new databases showing the high generalisation capability of the FV representation to successfully identify challenging unknown PAI species such as 3D Masks and Impersonation. 
    
    \item An extensive study over challenging cross-dataset scenarios including subject's ethnic, lighting conditions, and capture device variations. 
    
    \item In order to validate the detection capabilities of the proposal, a thorough evaluation compliant with the ISO/IEC 30107-3 standard on biometric PAD~\cite{ISO-IEC-30107-3-PAD-metrics-170227} is also carried out over well-established databases: CASIA Face Anti-Spoofing~\cite{Zhang-PAD-CASIA-ICB-2012}, REPLAY-ATTACK~\cite{Chingovska-PAD-RA-BIOSIG-2012}, REPLAY-MOBILE~\cite{Costa-Pazo-PAD-RM-BIOSIG-2016}, MSU-MFSD~\cite{Wen-PAD-MSU-ImgDistortion-TIFS-2015}, and SiW-M~\cite{Liu-PAD-DeepTree-CVPR-2019}.
    
    \item An extensive review of the state-of-the-art techniques employed for facial PAD. We mainly emphasis those methods focused on facial PAD generalisation. 
\end{itemize}

The remainder of this paper is organised as follows: a review of facial PAD methods is included in Sect.~\ref{sec:related_work}. Sect.~\ref{sec:proposal} presents the proposed PAD method. The experimental protocol allowed is explained in Sect.~\ref{sec:experiments}. The experimental results benchmarking the performance of our proposal with the top state-of-the-art techniques are discussed in Sect~\ref{sec:Results}. Finally, conclusions and future work directions are presented in Sect.~\ref{sec:conclusions}.

\section{Related Work}   
\label{sec:related_work}

The task of determining whether a sample stems from a live subject (i.e., it is a bona fide presentation, BP) or from an artificial replica (i.e., it is an attack presentation, AP) is a mandatory requirement which has received a lot of attention in the recent past. In order to mitigate the threats posed by attack presentations, a large number of PAD approaches have been proposed. They can be broadly classified as hardware- and software-based. 

\subsection{Hardware-based PAD techniques}

Hardware-based techniques include an additional sensor in the capture device in order to detect living characteristics of a human body such as intrinsic properties (e.g., reflectance~\cite{Kose-PAD-ReflectanceMask-DSP-2013,Wang-PAD-Multispectral-2013}), involuntary signals (e.g., thermal radiation~\cite{Sun-PAD-FaceThermo-2011}), or responses to external stimuli (e.g., motion estimation~\cite{Kollreider-PAD-MotionEstimation-2009}). In general, those methods report a high detection performance to identify specific PAI species. However, the inclusion of an extra sensor can significantly increase their development cost (e.g., a thermal sensor for an Iphone exceeds EUR 250\footnote{https://amz.run/44Mp}). In addition, given that such sensors are tailored for particular PAI species, their accuracy suffers a drastic decrease in the detection of unknown attacks~\cite{Galbally-PAD-FaceSurvey-2014}.   

\subsection{Software-based PAD techniques}

\subsubsection{Handcrafted-based methods}

Several software-based approaches have successfully spotted static PAI species (e.g., printed attacks) by analysing certain involuntary gestures: eye-blinking~\cite{Jee-PAD-EyeBlinking-2006,Pan-PAD-EyeBlinking-2007,Kollreider-PAD-EyeBlinking-2008,Patel-PAD-EyeBlinking-2016}, face and head gestures (e.g., nodding, smiling, looking in different directions)~\cite{Bigun-PAD-FaceTracking-2004,Ali-PAD-Gaze-2012,Tirunagari-PAD-2015}. However, they fail in the detection of PAI species such as printed attacks with cut eye regions and video replay attacks.

To compensate for such weaknesses, a large number of studies have addressed the PAD task by analysing texture properties. Handcrafted-based approaches usually employ processing tools such as: Fourier Spectrum to describe the global frequency of images~\cite{Li-PAD-FourierAnalysis-2004}, Gaussian filters to extract specific frequency information~\cite{Zhang-PAD-GaussianFilter-2012}, statistical models to detect image noise~\cite{Nguyen-PAD-FaceStatisticNoise-2019}, or traditional texture descriptors: LBP~\cite{Chingovska-PAD-RA-BIOSIG-2012,Xiong-UFacePAD-BTAS-2018,Peng-PAD-LBPEnsembleLearning-2020}, HOG~\cite{Agarwal-PAD-Multispectral-2017}, BSIF~\cite{Arashloo-PAD-BSIFfusion-2015}, LPQ~\cite{Raghavendra-PAD-MS-LPQ-2018}. Those techniques perform well for PAI species previously known a priori in training. However, their detection performance significantly decreases for unknown PAI species. 

\subsubsection{CNN-based methods}

The advances experienced by deep learning schemes in recent years and their great success in several computer vision tasks have led to the development of powerful architectures for PAD, which outperform the aforementioned handcrafted-based methods. In 2014, Yang \textit{et al.}~\cite{Yang-PAD-CNNApplicability-ArXiv-2014} fine-tuned ImageNet pre-trained CaffeNet~\cite{Jia-PAD-CaffeNet-2014} and VGG-face~\cite{Parkhi-VGG-Face-2015} models to distinguish a bona fide face sample from an attack presentation. Following this idea, Xu \textit{et al.}~\cite{Xu-PAD-LSTM-2015} combined Long Short-Term Memory (LSTM) units with CNNs to learn temporal features from face videos. The authors showed that the spatio-temporal features were helpful for facial PAD, thereby resulting in a reduction by half of the error rates reported by handcrafted feature baselines (5.93\% vs. 10.00\%). Keeping spatio-temporal features in mind, Gan~\textit{et al.}~\cite{Gan-PAD-3DCNN-ICMIP-2017} proposed a 3D CNN for facial PAD, which, unlike traditional 2D CNNs, extracts the temporal and spatial dimension
features from a frame sequence. Finally, Atoum \textit{et al.}~\cite{Atoum-PAD-depthCNN-IJCB-2017} also combined two-stream CNNs for extracting local features and depth estimation maps from facial images.  

In spite of the good detection rates achieved for these handcrafted- and deep learning-based approaches, they still face difficulties to identify PAIs when, $i)$ species employed in the fabrication of PAIs remain unknown in training (i.e., unknown PAI species) and $ii)$ samples in training and testing sets are acquired with different capture devices under different acquisition conditions (i.e., cross-dataset), thereby resulting in a poor generalisation.       

\subsubsection{Anomaly Detection-based methods}

\begin{figure*}[!t]
	\centering
	\includegraphics[width=\linewidth]{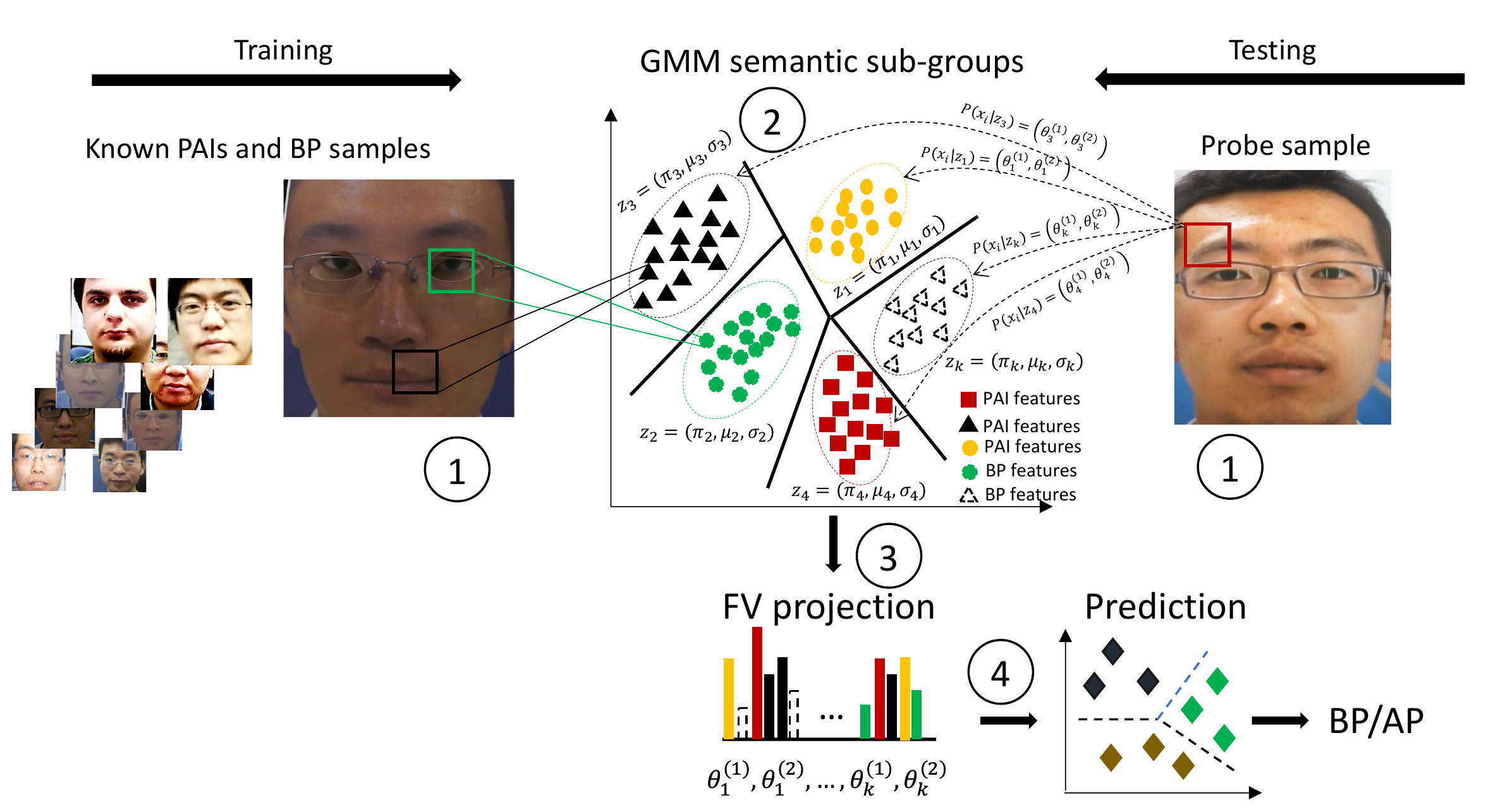}
	\caption{Face PAD approach overview which comprises three steps: $(1)$ local BSIF features are densely extracted, $(2)$ the feature distribution (i.e., semantic sub-groups) is subsequently learned by training an unsupervised Gaussian Mixture Model (GMM), $(3)$ the loglikelihood among the BSIF components and the parameters of the semantic sub-groups from the facial feature vector are computed; and $(4)$ the face representation is classified using a linear SVM.}
	\label{fig:overview}
\end{figure*} 

In order to tackle those shortcomings, several anomaly detection-based PAD methods have been proposed. In 2013, de Freitas Pereira \textit{et al.\ }\cite{FreitasPereira-FacePADRealWorld-ICB-2013} already reported poor generalisation capabilities to unknown attacks of state-of-the-art face PAD methods based on LBP and SVMs. In particular, the error rates increased by at least 100\%. Motivated by those findings, Arashloo \textit{et al.\ }\cite{Arashloo-AnomalyFacePAD-IEEEAccess-2017} experimented over several unknown attack scenarios and concluded that anomaly detection approaches trained only on bona fide data can reach a detection performance comparable to two-class classifiers. However, the results are reported only in terms of the area under the Receiving Operating Characteristic curve (AUC), thus lacking a proper quantitative analysis in line with the ISO/IEC 30107-3 standard on biometric PAD \cite{ISO-IEC-30107-3-PAD-metrics-170227}.

More recently, Nikisins \textit{et al.\ }\cite{Nikisins-AnomalyFacePAD-ICB-2018} showed how a one-class Gaussian Mixture Model (GMM) can outperform two-class classifiers depending on the PAI species included in the test set. The experimental results showed the generalisation capabilities of one-class classifiers with respect to two-class approaches for particular PAI species. Following the same anomaly detection paradigm, Xiong and AbdAlmageed studied in \cite{Xiong-UFacePAD-BTAS-2018} the detection performance of one-class SVMs and autoencoders in combination with LBP descriptors. In most of the scenarios tested, the detection rates increased with respect to common two-class classifiers. Liu \textit{et al.} also analysed in \cite{Liu-PAD-DeepTree-CVPR-2019} the performance of a Deep Tree Network (DTN) by clustering the PAI species into semantic sub-groups. Over a new database (SiW-M) comprising 13 PAI species and following a leave-one-out testing protocol, an average D-EER of 16\% is achieved, which is still above the state-of-the-art for known attacks. Finally, George and Marcel~\cite{George-PAD-OneClassRep-TIFS0-2020} also combined a one-class GMM with a Multi-Channel CNN (MCCNN), which are fed with face samples acquired by RGB, thermal, and infrared sensors. Despite the fact that the experimental evaluation over SiW-M dataset showed a performance improvement with respect to the DTN technique, its generalisation capability to detect unknown PAIs is still poor (i.e., a D-EER of 12.00\%).   

\subsubsection{Domain Adaptation-based methods}

Usually, acquisition properties such as facial appearance, pose, illumination, capture devices, and even subjects vary between datasets. In order to overcome poor cross-dataset generalisation issues, new PAD approaches have explored Domain Adaptation to transfer the knowledge learned from a source domain to a target domain~\cite{Ganin-DA-2015}. Yang \textit{et al.}~\cite{Yang-PAD-PersonSpecDA-TIFS-2015} proposed a subject domain adaptation method to synthesise virtual features by assuming that the relationship between BP and AP face samples on a same subject can be modelled with a linear transformation. Following this idea, Li~\textit{et~al.}~\cite{Li-PAD-MMD-DA-TIFS-2018} transformed knowledge learned from a labelled source domain to an unlabelled target domain by minimising the Maximum Mean Discrepancy~\cite{Long-DA-ResidualTransfer-2016} for facial PAD. De Freitas Pereira~\cite{deFreitas-FaceDomainAdaptation-2019} proposed a CNN-based method which builds a common feature space from face images, captured on different visual spectra domains, for improving face recognition. To transfer knowledge to the unlabelled target domain, Wang~\textit{et al.}~\cite{Wang-PAD-MultiDomDisRep-CVPR-2020,Wang-PAD-AdvDomainAdaptation-TIFS-2020} proposed an unsupervised domain adaptation with disentangled representation, which built a feature space shared for both source and target domains. Even if this common feature space appeared to be suitable to overcome cross-dataset issues, experimental results showed a poor detection performance over known attack scenarios (i.e., D-EERs of 3.20\%, 6.00\%, and 7.20\% for CASIA Face Anti-spoofing~\cite{Zhang-PAD-CASIA-ICB-2012}, MSU-MFSD~\cite{Wen-PAD-MSU-TIFS-2015}, and Rose-Youtu~\cite{Li-PAD-MMD-DA-TIFS-2018} databases, respectively).         

\subsubsection{Generative-based methods}

In the last decades, generative models have been in the vanguard of unsupervised learning. Techniques such as Gaussian Mixture Models (GMM)~\cite{Mclachlan-GMM-2004}, Boltzmann Machines (BMs)~\cite{Fahlman-BM-AAAI-1983}, Variational Autoencoders (VAEs)~\cite{Kingma-VAEs-ArXiv-2013}, and Generative Adversarial Networks (GANs)~\cite{Goodfellow-GANs-NIPS-2014} have been successfully applied in numerous computer vision~\cite{Krizhevsky-ImageNet-2012}, speech recognition and generation~\cite{Hinton-SpeechRecog-2012}, and natural language~\cite{Klein-NL-2003,Cotterell-VowelFormat-2018} tasks. Those algorithms try to capture the inner data probabilistic distribution to generate new similar data~\cite{Oussidi-PAD-GenerativeModels-ISCV-2018}. However, to the best of our knowledge, a rather limited number of works have been employed for PAD. Engelsma and Jain~\cite{Engelsma-GeneraFingSpoofDet-ICB-2019} fed several GANs with bona fide samples acquired by a RaspiReader fingerprint capture device. The experimental results for high-security threshold over unknown attacks showed a detection performance very sensitive to the training set.   


\section{Proposed Approach}
\label{sec:proposal}

We build our proposal keeping the generalisation capabilities of generative models in mind. Fig.~\ref{fig:overview} shows an overview of the proposed PAD subsystem, which consists of four main steps: (1) compact BSIF histograms are extracted from a face sample where the faces have been previously detected by the Viola and Jones method~\cite{Viola-FaceDetect-2004}; (2) semantic sub-groups are built by learning an unsupervised Gaussian Mixture Model (GMM) model from the aforementioned features; (3) the final descriptors are subsequently encoded by computing the differences of first- and second-order statistics with respect to the learned generative model parameters; and (4) a BP or AP decision is taken by a linear SVM. Linear SVMs are helpful since they perform well in high-dimensional spaces, avoid overfitting, and have good generalisation capabilities. 

\begin{figure}[!t]
	\centering
	\includegraphics[width=0.9\linewidth]{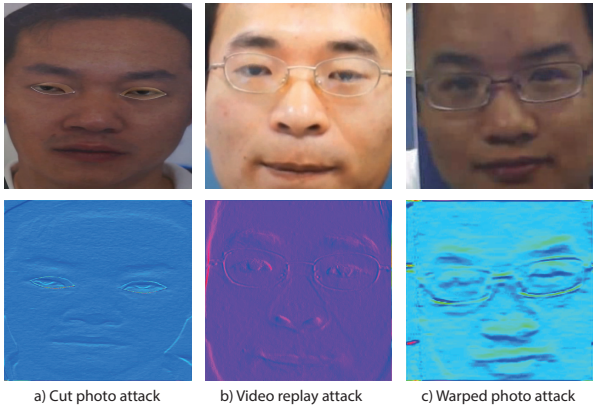}
	\caption{Visualisation of the artefacts on three PAI species after convolving the face image with a particular BSIF filter.}
	\label{fig:PAI_artefacts}
\end{figure} 

\subsection{Dense-BSIF Descriptors}
\label{sec:BSIF}

Most PAIs include properties which can be successfully detected by filtering the image with a particular kernel, as depicted in Fig.~\ref{fig:PAI_artefacts}. BSIF~\cite{Kannala-BISF-ICPR-2012} is a texture descriptor based on Independent Component Analysis (ICA)~\cite{Hyvarinen-NaturalImage-2009}, which uses a set of pre-trained filters to obtain a meaningful representation of the face data. More in details, given an image patch $X$ with size $l \times l$ pixels and a set of linear filters $W = \lbrace W_1, W_2, \dots, W_N \rbrace$ with the same size of $X$, the binarised response $b_i$ for $X$ can be computed as follows:

\begin{align}
b_i = \left\{ \begin{array}{cc} 
1 & \hspace{5mm} \sum_{n,m}W_i(n,m)X(n,m) > 0 \\
0 & \hspace{5mm} \text{otherwise} \\
\end{array} \right.
\end{align}

Once the binarised responses $b_i$ are computed for all filters $W_i: i = 1 \ldots N$, they are then stacked to form a bit string $\mathbf{b}$ with size $N$ for each pixel. Consequently, $\mathbf{b}$ is converted to a decimal value, and then a $2^N$ histogram for $X$ is yielded. In our experiments we employ 60 filter sets with different sizes $l = \{3, 5, 7, 9, 11, 13, 15, 17\}$ and number of filters $N= \{5, 6, 7, 8, 9, 10, 11, 12\}$, which were provided in~\cite{Kannala-BISF-ICPR-2012}. 

\begin{figure}[!t]
	\centering 
	\includegraphics[width=0.99\linewidth]{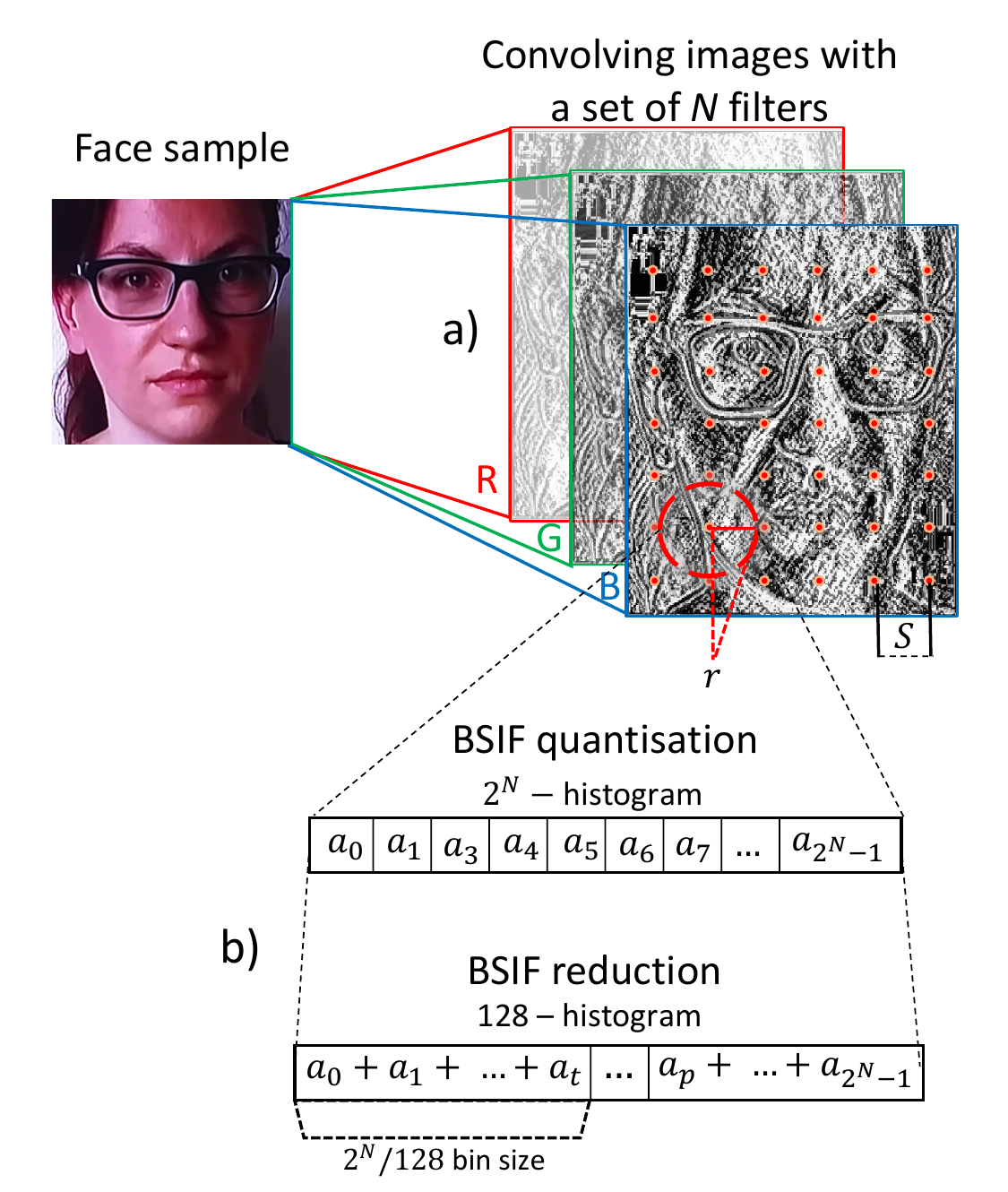}
	\caption{BSIF feature extraction on RGB images: a) BSIF histograms are densely computed at fixed points on a regular grid with a stride of $S$ pixels per channel, and b) the local $2^N$ BSIF histogram is then represented as a 128-component vector.}
	\label{fig:bsif_extraction}
\end{figure} 


Now, following the pipeline proposed in our preliminary study~\cite{Gonzalez-PAD-FVencForFacePAD-BIOSIG-2020}, the BSIF descriptors are densely extracted over a regular grid with a fixed stride $S$ of 3. Moreover, they are computed over four circular patches with different radii $r = \{ 4, 6, 8, 10\}$, as depicted in Fig.~\ref{fig:bsif_extraction}a), in order to capture local and global information of the artefacts produced in the creation of the PAIs. Therefore, each point in the grid poses four dense-BSIF histograms.

Since the BSIF features are sparse vectors as the number of linear filters $N$ increases, we also adopted the BSIF reduction strategy proposed in~\cite{Gonzalez-PAD-FVencForFacePAD-BIOSIG-2020}. Therefore, each $2^N$ BSIF histogram is reduced and represented as a 128-component vector by summing the elements for each sequential $2^N/128$ sub-set in the original histogram, as shown in Fig~\ref{fig:bsif_extraction}b). This representation, in turn, reduces both the BSIF computational cost and the storage requirements down to 12.5\% for N = 10 or 3.1\% for N = 12. 

\begin{table*}[!t]
	\processtable{A summary of databases included in our experimental protocol.\label{tab:db}}
{\begin{tabular*}{42pc}{@{\extracolsep{\fill}}cclll@{}}\toprule
DB & \#Samples & Capture device & Capture conditions	& 	PAI species\\
\midrule
\multirow{4}{*}{CASIA FASD} & \multirow{4}{*}{600} 	   & Low-quality USB camera      							& \multirow{3}{*}{Natural scenes}	&	Warped photo or Printed attacks, \\ 
	                                         &         & Normal-quality USB camera 								& 									&	\multirow{2}{*}{Cut photo, Video replay}	\\ 
                                             &         & High-quality Sony NEX-5 camera 						&  									&		\\
\midrule
\multirow{2}{*}{REPLAY-ATTACK} & \multirow{2}{*}{1,200} & \multirow{2}{*}{Low-quality 13-inch MacBook webcam} 	&	Controlled,						& \multirow{2}{*}{Printed, Photo replay, Video replay} \\
                                             &          & 				 										& 	adverse scenes					&			\\ 
\midrule
\multirow{4}{*}{REPLAY-MOBILE} & \multirow{4}{*}{1,190}& \multirow{2}{*}{High-quality IPad Mini 2}      		&	Controlled, adverse				& \multirow{4}{*}{Printed, Photo replay, Video replay} \\
	                                         &         & 					 									& 	direct sunlight, 				&\\
                                             &         & 	High-quality LG G4						  			& 	lateral sunlight,				&\\ 
											 &         & 							 							& 	diffuse and complex backgrounds	&\\ 
\midrule
\multirow{3}{*}{MSU-MFSD} 		& \multirow{3}{*}{440} & \multirow{2}{*}{Low-quality 13-inch MacBook webcam}  	& 	\multirow{3}{*}{Natural scenes}	& \multirow{3}{*}{Printed, Video replay}  \\
											&          & 								 		   				&									& \\
											&          & Low-Quality Google Nexus 5 camera						& 									& \\
\midrule
\multirow{4}{*}{SiW-M} 			& \multirow{4}{*}{968} & \multirow{2}{*}{High-quality Logitech C920 webcam}		& 	\multirow{2}{*}{Controlled,} 	& Printed, Video replay, Half mask, Silicone mask, \\
											&          & 						 		   						& 									& Transparent, Papercraft, Mannequin, Obfuscation, \\
											&          & High-quality Canon EOS T6								& 	adverse scenes					& Impersonation, Cosmetic, Funny Eye, \\
											&          & 						 		   						& 									& Paper Glasses, Partial Paper\\
			\bottomrule
	\end{tabular*}}{}
\end{table*}

\subsection{Fisher Vector Encoding}
\label{sec:FV}

The FV encoding derives a kernel from the parameters of generative model parameters (i.e., GMM~\cite{Sanchez-ImgClassificationFV-2013} for our work). This representation characterises how the distribution of a set of local descriptors, extracted from unknown PAI species, differs from the distribution of known APs and BPs, which is previously learned by a generative model. Therefore, the final transformed features are more robust to new samples, which may stem from unknown scenarios and thus differ from the samples used for training, as shown in a preliminary evaluation in~\cite{Gonzalez-PAD-FVencForFacePAD-BIOSIG-2020}.

In this article, we follow the idea in~\cite{Perronnin-ImprovingFV-ECCV-2010} and train a GMM with diagonal covariances from a set of local features (i.e., compact dense-BSIF descriptors). In particular, a GMM on $K$-components, which is represented by their mixture weights~($\pi_k$), means~($\mu_k$), and covariance matrices~($\sigma_k$), with $k = 1, \dots, K$, allows discovering semantic sub-groups from known PAIs and BP samples, which could successfully enhance the detection of unknown attacks. Given that there is a high correlation between the three RGB colour components~\cite{Boulkenafet-PAD-ColorGeneralisation-2018}, the semantic sub-groups are built by decorrelating firstly the current local descriptors with Principal Component Analysis (PCA)~\cite{Jegou-VLAD-PAMI-2012}. This, in turn, reduces their size to $d = 64$ components while retaining 95\% of the variance. Then, the FV representation captures the average first-order and second-order statistic differences between the local features and each semantic sub-group previously learned by the GMM~\cite{Simonyan-FV-BMVC-2013}. 

Let $\mathbf{X}$ be a local descriptor of size $d$ and $G_K=\{(\pi_k, \mu_k, \sigma_k): k=1\ldots K\}$ a set of $K$ semantic sub-groups learned by the GMM. The FV representation for $\mathbf{X}$ is defined as the conditional probability:

\begin{align}
	FV_X = 	& 	P(X| \mu_k, \sigma_k) 
\end{align}

By applying Bayesian properties, we can rewrite the previous equation as:

\begin{align}
	\phi_k^{1} = & \frac{1}{N\sqrt{\pi_k}}\sum_{i = 1}^d \alpha_i (k) \left(\frac{X_i - \mu_k}{\sigma_k}\right), \\	
	\phi_k^{2} = & \frac{1}{N\sqrt{2\pi_k}}\sum_{i = 1}^d \alpha_i (k) \left(\frac{(X_i - \mu_k)^2}{\sigma_k^2} - 1\right), 
\end{align}

\noindent where $\alpha_i(k)$ is the soft assignment weight of the $i$-th feature $x_i$ to the $k$-th Gaussian. Finally, the FV representation that defines a facial image is obtained by stacking the differences $\phi_G~=~\left[ \phi_1^{1}, \phi_1^{2}, \cdots,  \phi_K^{1}, \phi_K^{2} \right]$, thereby resulting a $2 \cdot d \cdot K = 2 \cdot 64 \cdot K$ sized vector.


\subsection{Classification}
\label{sec:classif}

For the BP vs. AP decision, a linear SVM has been employed. According to~\cite{Hsu-SVMPracticalGuide-2003}, when the feature's dimensionality is so big in comparison with the number of instances employed for training, a non-linear mapping does not improve the performance. Therefore, the use of a linear kernel would be good enough to achieve a high classification accuracy. 

In order to find the optimal hyperplane separating the bona fide from the attack presentations, the optimisation algorithm bounds the loss from below. Therefore, we have trained a linear SVM as follows: The SVM labels the bona fide samples as +1 and the presentation attacks as -1, thereby yielding the corresponding $W$ (weights) and $\mathbf{b}$ (bias) classifier parameters. Then, given a FV encoding $\mathbf{x}$, the final score $s_{\mathbf{x}}$, which estimates the class of the sample at hand, is computed as the confidence of such decision (i.e., the absolute value of the score is the distance to the hyperplane):

\begin{eqnarray}
s_{\mathbf{x}} &= W \cdot \mathbf{x} + \mathbf{b}
\end{eqnarray}

\section{Experimental Setup}
\label{sec:experiments}

The experimental evaluation aims to address the following goals: $i)$ analyse the impact of different BSIF filter configurations in terms of the number of filters and filter's size on the PAD performance, $ii)$ study the detection performance for different colour spaces (i.e., RGB, HSV, and $\mathrm{YC_{b}C_{r}}$), and $iii)$ benchmark the detection performance of our PAD approach with the top state-of-the-art for known and unknown attacks. Keeping these goals in mind, we define three different scenarios: 

\begin{itemize}

\item \textit{Known-attacks}, which includes an analysis of all PAI species. In all cases, PAI species for testing are also included in the training set, as described in \cite{Zhang-PAD-CASIA-ICB-2012}. On this scheme, we carry out the parameter optimisation as well as the study of the three colour spaces which serve as a baseline for the remaining scenarios. 

\vspace*{0.1cm}

\item \textit{Unknown PAI species}, in which the PAI species used for testing are not incorporated in the training set. We use the leave-one-out testing protocol explained in~\cite{Arashloo-AnomalyFacePAD-IEEEAccess-2017}

\vspace*{0.1cm}

\item \textit{Cross-database}, in which the datasets employed for testing are different from the databases used for training. Both datasets contain the same PAI species to ensure that the performance degradation is due to the dataset change and not to the unknown PAI species.
\end{itemize}

\subsection{Databases}
\label{sec:datasets}

In order to reach our goals, the experimental evaluation was conducted over five well-established databases, which are summarised in Tab.~\ref{tab:db}:  


\begin{itemize}

    \item[\textbf{CASIA Face Anti-Spoofing database}~\cite{Zhang-PAD-CASIA-ICB-2012}] contains 600 short videos of bona fide and attack presentations stemming from 50 different subjects and acquired under different conditions. The dataset comprises three PAI species: $i)$ warped photo attacks or printed attacks, in which the attackers place their face behind the hard copies of high-resolution digital photographs, $ii)$ cut photo attacks, the face of the attacker is placed behind the hard copies of photos, where eyes have been cut out, and $iii)$ video replay attacks, where attackers replay face videos using iPads.
    
    \vspace*{0.1cm}

    \item[\textbf{REPLAY-ATTACK}~\cite{Chingovska-PAD-RA-BIOSIG-2012}] consists of 1200 short videos (around 10 seconds in mov format) of both bona fide and attack presentations of 50 different subjects, acquired with a 320 $\times$ 240 low-resolution webcam of a 13-inch MacBook Laptop. The video samples were recorded under two different conditions: $i)$ controlled, with uniform background and artificial lighting, and $ii)$ adverse, with natural illumination and non-uniform background. In addition, this database comprises three PAI species: printed attacks, photo replay attacks (i.e., a mimic photo is replayed by a smartphone to the capture device), and video replay attacks. 
    
    \vspace*{0.1cm}
    
    \item[\textbf{REPLAY-MOBILE}~\cite{Costa-Pazo-PAD-RM-BIOSIG-2016}] comprises 1190 video clips of printed attacks, photo replay attacks, and video replay attacks of 40 subjects under different lighting conditions. Those videos were recorded with two smart capture devices: an iPad Mini2 and a LG-G4 smartphone, thereby allowing the evaluation of PAD approaches for the mobile scenario.
    
    \vspace*{0.1cm}
    
    \item[\textbf{MSU-MFSD}~\cite{Wen-PAD-MSU-TIFS-2015}] contains 440 video clips of photo replay attacks and video replay attacks of 35 subjects. Those PAI species were acquired with two camera types: MacBook Air 13-inch and front-camera in the Google Nexus 5 smartphone. The MSU-MFSD database comprises two particular scenarios: $i)$ a mobile phone is used to capture both bona fide presentations and presentation attacks, simulating the application of mobile phone unlock, and $ii)$ the printed photos used for attacks are generated with a state-of-the-art colour printer on larger sized paper.
    
    \vspace*{0.1cm}
    
    \item[\textbf{SiW-M}~\cite{Liu-PAD-DeepTree-CVPR-2019}] consists of 968 videos of 13 PAI species including challenging attacks such as silicone masks, obfuscation, and cosmetic makeup, among others. 660 bona fide videos from 493 subjects are also included in the dataset. Those subjects are diverse in ethnicity and age, and the videos were collected in 3 sessions: $i)$ a room environment where the subjects were recorded with few variations such as pose, lighting and expression; $ii)$ a different and so larger room where the subjects were recorded with lighting and expression variations; and $iii)$ a mobile phone mode where the subjects are moving while the phone camera is recording. Extreme pose angles and lighting conditions are also introduced.
\end{itemize}


\subsection{Evaluation Metrics}
\label{sec:metrics}

Finally, all results are analysed and reported in compliance with the metrics defined in the international standard ISO/IEC 30107-3~\cite{ISO-IEC-30107-3-PAD-metrics-170227} for biometric PAD:


\begin{itemize}
    \item Attack Presentation Classification Error Rate (APCER), which is defined as the proportion of attack presentations wrongly classified as bona fide presentations.
    
    \vspace*{0.1cm}

    \item Bona Fide Presentation Classification Error Rate (BPCER), which is the proportion of bona fide presentations misclassified as attack presentations.
\end{itemize}   

Based on these metrics, we report: $i)$ the Detection Error Trade-off (DET) curves between APCER and BPCER; $ii)$ the BPCERs observed at different APCER values or security thresholds such as 10\% (BPCER10), 5\% (BPCER20), and 1\% (BPCER100), respectively; and $iii)$ the Detection Equal Error Rate (D-EER), which is defined as the error rate value at the operating point where APCER~=~BPCER.

\section{Experimental Results}
\label{sec:Results}

\begin{table}[!t]
	\centering
	\processtable{Detection performance, in terms of D-EER (\%), of our proposed approach for different $K$ values. The best result is highlighted in bold. \label{tab:optimisation_k}}
	{\vspace*{0.1cm}
	\begin{tabular}{l c c c} 
	\toprule
	\backslashbox{DB}{$K$} 							& 		256				&  			512				& 		1024			\\
	\midrule
					CASIA-FASD    					&	2.02 $\pm$ 0.93		&   	1.95 $\pm$ 0.79		&  	1.79 $\pm$ 0.82		\\			
	
					REPLAY-ATTACK    				&  	0.00 $\pm$ 0.00		&		0.00 $\pm$ 0.00		&	0.00 $\pm$ 0.00   	\\
	
					REPLAY-MOBILE      				& 	0.01 $\pm$ 0.03		&		0.00 $\pm$ 0.00 	&   0.00 $\pm$ 0.00	    \\
					
					MSU-MFSD      					& 	0.02 $\pm$ 0.09		&		0.01 $\pm$ 0.08	 	&   0.01 $\pm$ 0.08     \\
	\midrule
					Avg.							&		0.51			&			0.49			&	 \textbf{0.45}		\\
	\bottomrule
	\end{tabular}}{}
\end{table}

\subsection{Known Attacks}
\label{sec:intra_attacks}

\subsubsection{Effects of the number of semantic sub-groups}
\label{sec:intra_attacks_semantic}

First, we need to find the optimal configuration of our proposed method in terms of the key parameters: the filter size $l$, the number of BSIF filters $N$, and the number of semantic sub-groups $K$. Following the overall protocol provided by the datasets~\cite{Zhang-PAD-CASIA-ICB-2012,Chingovska-PAD-RA-BIOSIG-2012,Costa-Pazo-PAD-RM-BIOSIG-2016,Wen-PAD-MSU-TIFS-2015}, we compute the D-EER for each of sixty filter configurations (i.e., one error rate for each filter set employed by our dense-BSIF) and report in Tab.~\ref{tab:optimisation_k} the mean and standard deviation for each fixed $K$ value. In all experiments, we tested the value range $K~\in$~\{256, 512, 1024\}: $K$ values greater than 1024 would result in large feature vectors that are not suitable for real-time applications and hence are not considered in this work. As it may be observed, the detection performance of our method increases with $K$: a D-EER of 0.45\% on average is achieved for $K$ = 1024. 
Therefore, this $K$ value will be considered for the remaining experiments. In addition, we may observe that the standard deviation is below 1.0\% in all datasets, hence indicating that a statistically meaningful representation of face data can be obtained using different BSIF filters, regardless of the values chosen for $N$ and $l$. 

It should be noted that there is a high difference between the error rates attained for CASIA and the ones achieved for the remaining datasets. Specifically, the D-EERs for CASIA are up to 20 times greater than the ones reported for other databases. We think that this divergence is mainly given by the image resolution employed for training and testing our approach. Whereas the REPLAY-ATTACK, REPLAY-MOBILE, and MSU-MSFD databases consist of images acquired with fixed low or high-resolution capture devices respectively, face images in CASIA were obtained with a mix of low-, medium-, and high-resolution capture devices. Motivated by that fact, we analysed in~\cite{GonzalezSoler-PAD-ImgResolutionSens-NISK-2020} the sensitivity of several PAD approaches to images with varying resolutions. As a result of that work, we showed that both deep learning- and handcrafted-based PAD techniques suffered a high-performance deterioration when they were trained with datasets having images of different resolutions: D-EERs achieved for CASIA were increased by up to 20\%. We do confirm that our approach is also affected by the image quality varying which should be carefully analysed for more challenging scenarios.        

\begin{table}[!t]
	\centering
	\processtable{Detection performance in terms of D-EER (\%) of our proposed approach for the best performing $K = 1024$. \label{tab:optimisation_colour_space}}
	{
	\begin{tabular}{l c c c} 
	\toprule
	\backslashbox{DB}{Colour}		 							& 		RGB				&  		HSV				& 	\textbf{$\mathrm{YC_{b}C_{r}}$}		\\
	\midrule
					CASIA-FASD    								&	1.79 $\pm$ 0.82		&  2.35 $\pm$ 1.07		&  	2.20 $\pm$ 1.05		\\			
	
					REPLAY-ATTACK    							&  	0.00 $\pm$ 0.00		&  0.02 $\pm$ 0.05		&	0.10 $\pm$ 0.26   	\\
	
					REPLAY-MOBILE      							& 	0.00 $\pm$ 0.00		&  0.00 $\pm$ 0.00 		&   0.12 $\pm$ 0.26	    \\

					MSU-MFSD      								& 	0.01 $\pm$ 0.08		&  0.08 $\pm$ 0.35 		&   0.35 $\pm$ 0.81	    \\
	\midrule
					Avg.										&	\textbf{0.45}		&		0.79			&		  0.69			\\
	\bottomrule
	\end{tabular}}{}
\end{table}

\subsubsection{Colour space analysis}
\label{sec:intra_attacks_rgb}

According to Boulkenafet \textit{et al.}~\cite{Boulkenafet-PAD-ColorGeneralisation-2018}, the RGB colour space has limited discriminative power for face PAD due to the high correlation between the three colour components. In contrast, HSV and $\mathrm{YC_{b}C_{r}}$ are based on the separation of the luminance and chrominance components, thereby providing additional information for learning more discriminative features. Based on that observation, we evaluate in Tab.~\ref{tab:optimisation_colour_space} the detection performance of our proposed approach for the three aforementioned colour spaces. As it can be seen and contrary to the conclusions drawn in~\cite{Boulkenafet-PAD-ColorGeneralisation-2018}, RGB appears to be the colour space including the most discriminative features for facial PAD, thereby resulting, on average, in a D-EER of 0.45\%. However, taking a closer look, we can observe that the three colour spaces report similar error rates in three out of four datasets (i.e., REPLAY-ATTACK, REPLAY-MOBILE, and MSU-MFSD): mean D-EERs of 0.003\%, 0.03\%, and 0.19\% are achieved by RGB, HSV, and $\mathrm{YC_{b}C_{r}}$ respectively.

In order to validate the detection performance achieved by our proposed method using current colour spaces, we select the non-parametric Mann-whitney test~\cite{Lowry-StatConcept-2014} with a 95\% of confidence to verify the statistical significance of the sixty error rates reported by different colour spaces. To that end, we define the null hypothesis $H_0$ and alternative hypothesis $H_1$ as:

\begin{itemize}
    \item $H_0$: two colour spaces provide the same discriminative information for face PAD.
    \item $H_1$: two colour spaces do not provide the same discriminative information for face PAD.
\end{itemize}

Then, an all-against-all comparison per dataset is performed. As a result of this test, we do confirm that the RGB only provides the most discriminative information for one out of four databases: error rates attained by the RGB claim to be statistically higher than the ones reported by the other colour spaces for the CASIA database. In contrast, for the three remaining databases (i.e., REPLAY-ATTACK, REPLAY-MOBILE, and MSU-MFSD), Mann-whitney results state that the three colour spaces include the same discriminative information, thereby confirming their similar detection performances reported in Tab.~\ref{tab:optimisation_colour_space}. The reason for this difference with respect to~\cite{Boulkenafet-PAD-ColorGeneralisation-2018} is that we carried out a feature decorrelation with PCA before finding the semantic sub-groups, thereby leading to the detection of similar features for the three colour spaces.    


\begin{table}[!t]
	\scriptsize	
	\centering
	\processtable{Benchmark with state-of-the-art in terms of D-EER (\%) for the \textit{Known-attacks} scenario using $K = 1024$ on RGB.\label{tab:benchmark_sota}}
	{
		\resizebox{\linewidth}{!}{
		\begin{threeparttable}
		\begin{tabular}{r|cccc} \toprule
								Method   							 		    & 	CASIA 				& 			RA	 			 &		MSU				&		 RM					\\ \midrule
			BSIF-SVM~\cite{Raghavendra-PAD-PadForIrisAndFace-EUSPICO-2014}		&	 10.2		  		&			 -		  		 &						&		  -		  				\\	
			MBSIF-TOP~\cite{Arashloo-PAD-BSIFfusion-2015}		   				&    7.20		  		&			 -		   		 &						&		  -		  				\\
			CSURF + FV~\cite{Boulkenafet-PAD-SURFfV-2016}						&    2.80		  		&			0.10 		   	 &		2.20			&		  -		  				\\
			Texture fusion~\cite{Boulkenafet-PAD-ColorGeneralisation-2018}		&    4.60    	  		&			1.20			 &		1.50			&		  -						\\
			Depth CNNs~\cite{Atoum-PAD-depthCNN-IJCB-2017}				   		&    2.67	      		&			0.72	 		 &	0.35 $\pm$ 0.19		&	 	  -					     \\
			ResNet-15-3D~\cite{Guo-PAD-3DvirtualSynthesis-ICB-2019}				&	 2.22	   	  		&		  	0.25			 &		 -				&		  -						 \\
			FaceSpoofBuster~\cite{Bresan-PAD-FaceSpoofBuster-2019}	   			&	 3.88	 	  		&			5.50	 		 &		 -				&		  -						  \\
			shallowCNN-LE~\cite{Qu-PAD-ShallowCNN2019} 							&	 4.00	      		&			3.70	  		 &	    8.41			&		  -						  \\
			DR-UDA~\cite{Wang-PAD-AdvDomainAdaptation-TIFS-2020}	 			&	 3.30		  		&	1.30\tnote{$\ddagger$}	 &		6.30			&		  -					 	  \\ 
			SPMT + SSD~\cite{Song-PAD-discriminative-PR-2019}   				&	 0.04		  		&			0.03	   		 &	 	 -				&		  -						  \\
			DeepPixBiS~\cite{George-PAD-DeepPixBis-2019}   						&	 -			  		&			 -		   		 &		 -				&		 0.00					  \\
			WeightedAvg.~\cite{Fatemifar-PAD-WeightedAvg-ICB-2019}   			&	 -			  		&			1.43		   	 &		 -				&		 9.95					  \\
			   HR-CNN~\cite{Muhammad-PAD-HybridResidual-ICB-2019}	 			&	 0.02		  		&			 -			     &		0.04			&		  -					 	  \\ \midrule
		   Our Method   	  													& 1.79 $\pm$ 0.82 		&    0.00 $\pm$ 0.00		 &	0.01 $\pm$ 0.08		&	0.00 $\pm$ 0.00				 \\
		   Best D-EER\tnote{$\dagger$}			   	  							& 0.37					&	 0.00 					 &	0.00 				&	0.00 						\\
		   BPCER @ APCER = 1.0\%  	  											& 0.00 					&	 0.00		 			 &	0.00				&	0.00				 		 \\
		\bottomrule
		\end{tabular}
		\begin{tablenotes}\footnotesize
			\item[$\dagger$] The best D-EER as well as the BPCER @ APCER = 1.0\% per dataset are attained for $N$ = 10 filters of size $l$ = 9.
			\item[$\ddagger$] Half Total Error Rates (HTER) reported in~\cite{Wang-PAD-AdvDomainAdaptation-TIFS-2020}  
		\end{tablenotes}
	\end{threeparttable}}}{}
\end{table} 

\begin{table*}[!t]
		\begin{center}
			\processtable{Benchmark with the state-of-the-art in terms of the AUC (\%) for $K = 1024$ and RGB over \textit{traditional unknown PAI species}. The best results per PAI species are highlighted in bold.\label{tab:single_training_cross_attacks}} 
			{
			\resizebox{\textwidth}{!}{
			\setlength\tabcolsep{3.0pt}
			\begin{threeparttable}
			\begin{tabular}{r |ccc|ccc|ccc|ccc} \toprule
																	 & 			\multicolumn{3}{c|}{CASIA} 	  	   		   & 		\multicolumn{3}{c|}{REPLAY-ATTACK}			& 			\multicolumn{3}{c|}{MSU-MFSD}					& 				\multicolumn{3}{c}{REPLAY-MOBILE}			\\ 
																	 & 		 Cut 		&      Warped    &     Video  	   &     Digital 	 &     Printed    & 	Video 	 	&  	Printed 		&  		HR Video  	& 	Mobile Video    &  	Digital 		&  		Printed  	& 		Video		\\ \midrule
	OC-SVM\_RGB+BSIF \cite{Arashloo-AnomalyFacePAD-IEEEAccess-2017} &  	60.70   	&		95.90  	 &      70.70  	   &		88.10	 & 		 73.70	  & 	 84.30		& 	 64.80			&  		87.40 		& 	  74.70			& 		-			&  			-	 	& 		  -			\\
					 NN+LBP \cite{Xiong-UFacePAD-BTAS-2018} 		 &   	88.40  		&		79.90  	 &      94.20	   &		95.20	 & 		 78.90	  & 	 99.80		& 	 50.60			&  		99.90	 	& 	  93.50			& 		-			&  			-	 	& 		  -			\\ 
						DTN \cite{Liu-PAD-DeepTree-CVPR-2019} 	     &   	97.30  		&		97.50  	 &      90.00	   &		99.90	 & 		 99.60	  & 	 99.90		& 	 81.60			&  		99.90	 	& 	  97.50			& 		-			&  			-	 	& 		  -			\\
						CDCN \cite{Yu-PAD-CDC-CVPR-2020} 	     	 &  \textbf{99.90}  & \textbf{99.80} &      98.48	   &		99.43	 & 		 99.92	  & \textbf{100}	& 	 70.82			&  \textbf{100}	 	& 	  99.99			& 		-			&  			-	 	& 		  -			\\ \midrule
					 our proposal (AUC)  	    					 &      99.6	 	& 		97.9  	 & \textbf{99.9}   &	\textbf{100} & \textbf{100}  &  \textbf{100}	& \textbf{99.32}	&  \textbf{100}	 	& \textbf{100} 		& \textbf{100}		&  \textbf{100}	 	& \textbf{100} 		\\
					 our proposal (D-EER)\tnote{$\star$}   	 		 & 		3.33		& 		6.67	 & 		2.22 	   & 		0.00 	 & 		 0.00	  & 	 0.00 		& 	 1.96			&  		0.00		& 	  0.00	 	    & 	   0.00			&  		0.00		& 		0.00	 	\\
					 our proposal (mean D-EER)   	 			     & 4.11 $\pm$ 1.99	& 6.15 $\pm$ 2.42& 1.37 $\pm$ 1.60 & 0.00 $\pm$ 0.00 & 1.35 $\pm$ 1.73& 0.00 $\pm$ 0.00	& 6.64 $\pm$ 4.62	&  0.00 $\pm$ 0.00	& 0.11 $\pm$ 0.44	& 0.00 $\pm$ 0.00	&  0.34 $\pm$ 0.63	& 0.02 $\pm$ 0.12	 \\
					 \bottomrule
			\end{tabular}
			\begin{tablenotes}\footnotesize
				\item[$\star$] The D-EER and AUC values per dataset are reported for $N$ = 10 filters of size $l$ = 9.
			\end{tablenotes}
		\end{threeparttable}}}{}
		\end{center}
	\end{table*}

	\begin{figure*}[!t]
		\centering
		\includegraphics[width=0.80\linewidth]{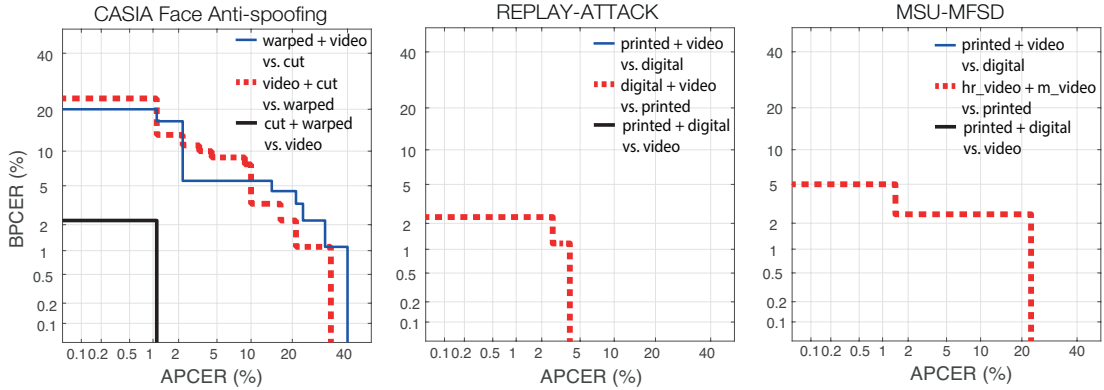}
		\caption{\textit{Traditional unknown PAI species} DET curves over the leave-one-out protocol for the CASIA, REPLAY-ATTACK, and MSU-MFSD databases. The REPLAY-MOBILE database reports a remarkable BPCER = 0.0\% for any APCER.}
		\label{fig:cross_attacks_det}
	\end{figure*} 

\subsubsection{Benchmark with the state-of-the-art}
\label{sec:intra_attacks_semantic}

Finally, we benchmark in Tab.~\ref{tab:benchmark_sota} our approach with the top state-of-the-art PAD techniques for the best performing colour space and $K$ value (i.e., RGB and $K$ =  1024). Firstly, it can be observed how a baseline implementation based on BSIF and SVMs in~\cite{Arashloo-PAD-BSIFfusion-2015}, where extracted features have not been transformed with the FV technique, reports a D-EER of 10.21\%, in contrast to the best error rate achieved in this work for CASIA (i.e., 0.37\%). 

On the other hand, it may be also observed that the FV representation does not produce a reliable detection performance: it depends on a good feature extractor for the specific data domain. Specifically, the combination of FV and SURF descriptors, which has shown a remarkable detection performance for fingerprint PAD~\cite{Gonzalez-PAD-MaterialImpact-ICB-2019}, achieves D-EERs of 2.80\% and 0.10\% for CASIA and REPLAY-ATTACK respectively, which are still far away from the ones reported in this work. Therefore, we can conclude that the use of compact dense-BSIF descriptors presents a clear advantage for facial PAD with respect to gradient-based features such as SURF. 

In addition, the texture fusion approach in~\cite{Boulkenafet-PAD-ColorGeneralisation-2018} can be also outperformed by a relative 96\% and 100\% respectively, depending on the testing database. Among the next five deep learning-based techniques, the lowest D-EER reported are 2.22\% and 0.25\%, which are also twelve and twenty-five times worse than our best results for CASIA and REPLAY-ATTACK databases, respectively. In contrast, the last two approaches analysed~\cite{Muhammad-PAD-HybridResidual-ICB-2019,Song-PAD-discriminative-PR-2019} outperform our technique by one order of magnitude for the CASIA database. However, our best result is three-time better than the ones reported in~\cite{Song-PAD-discriminative-PR-2019} for REPLAY-ATTACK (i.e., 0.00\% vs 0.03\%). It should be noted that the authors of those works admit that their PAD approaches are time-consuming methods. Very low computational complexity is an additional advantage of our approach, which needs about 0.7 seconds per classification attempt, thereby making it suitable for real-time applications. Finally, we can observe that for a high-security threshold (i.e., APCER = 1.0\%), our proposed method reports a remarkable BPCER of 0.0\% for all databases: only one in 100 attack presentation attempts are accepted while zero BPs are rejected by our algorithm when PAI species and capture devices employed in the PAI acquisition are known a priori. 

\subsection{Unknown PAI species}
\label{sec:unk_PAI}

As it was mentioned in Sect.~\ref{sec:intro}, one of the main goals of this work is to address the detection of unknown attacks. In particular, we tackle the challenging scenario where PAI species remain unknown in the training set of PAD techniques. To that end, two sets of experiments are carried out over the five selected databases following the leave-one-out protocol described in~\cite{Arashloo-AnomalyFacePAD-IEEEAccess-2017}: one PAI specie is evaluated while the remaining PAI species are included in the training set. 

\subsubsection{Generalisation across traditional unknown PAI species}

In the first set of experiments, we evaluate the feasibility of our proposed method to detect unknown attacks over traditional PAI species (i.e., printed attacks, cut photo attacks, and photo and video replay attacks). The corresponding results are reported in Tab.~\ref{tab:single_training_cross_attacks}. Firstly, it should be noted that error rates for each particular unknown PAI species in CASIA are multiplied by a factor of 2.17\% on average with respect to the corresponding D-EERs reported in Tab.~\ref{tab:benchmark_sota} (i.e., 3.88\% vs. 1.79\%). In contrast, the D-EERs for the remaining datasets are comparable with their corresponding error rates for the known attack scenario. These observations confirm the sensitivity of our approach to training with datasets having images of varying resolutions. 

Regarding the MSU-MFSD database, our method suffers a performance deterioration for printed attacks. Whereas both HR video and mobile video attacks report on average a D-EER of 0.0\%, printed attacks attain an average error rate of 6.64\%, hence indicating that BSIF texture features of the latter are as close to BP semantic sub-groups as the semantic sub-groups defined from video replay attacks. This, in turn, states that the detection performance over unknown PAI species depends on a reliable known PAI species selection for training. Due to the lack of a proper quantitative analysis of the top state-of-the-art methods in compliance with the ISO/IEC 30107-3 standard on biometric PAD~\cite{ISO-IEC-30107-3-PAD-metrics-170227}, we establish a benchmark in terms of Area Under the Curve (AUC). In spite of the previous shortcomings, we can note that for a fixed filter configuration (i.e., $N$ = 10 filters of size $l$ = 9 pixels), our approach achieves current state-of-the-art results for all datasets, thereby resulting in an AUC closed to 100\%. 

Finally, a high detection performance of our method can be perceived in Fig.~\ref{fig:cross_attacks_det}: a BPCER in the range of 0.0\% - 17\% for any APCER~$\ge$~1.0\% confirms the soundness of the common feature space defined by FV to separate an unknown AP from a BP attempt.   

\subsubsection{Generalisation across challenging unknown PAI species}

\begin{figure}[!t]
	\centering
	\includegraphics[width=0.9\linewidth]{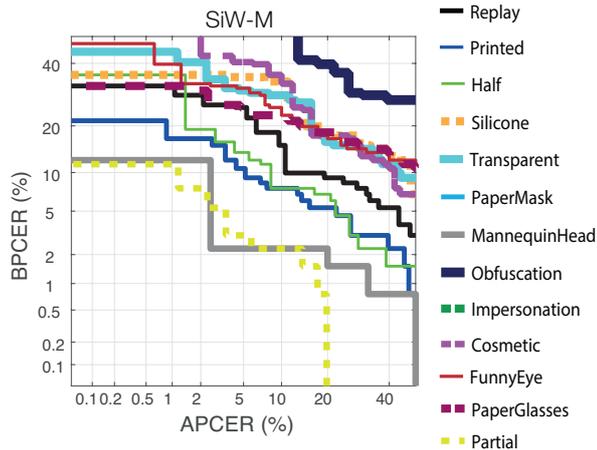}
	\caption{\textit{Challenging unknown PAI species} DET curves over the leave-one-out protocol for the SiW-M database on the best BSIF filter configuration.}
	\label{fig:challenging_unk_PAIs_det}
\end{figure}

In the second set of experiments, we evaluate challenging unknown PAI species such as 3D Masks (i.e., Silicone masks, Transparent masks, and Mannequin Head) and Makeup (obfuscation, impersonation, and cosmetic) in the SiW-M database following the leave-one-out protocol: twelve PAI species are employed for training and the remaining thirteenth PAI species is used for testing. It is worth point out that there is no overlap between training and test subjects. Tab.~\ref{tab:unknowPAI_SIW} reports the D-EER for $N$ = 10 BSIF filters of size $l$ = 9 and the best BSIF performing filter configurations. The corresponding DET curves for the latter are depicted in Fig.~\ref{fig:challenging_unk_PAIs_det}. As it may be observed, our best filter performing-based FV representation reports an improvement with respect to the results attained by state-of-the-art methods, thereby yielding a D-EER of 11.44\% and a standard deviation of 8.73\%. We can also note that this approach attains the top state-of-the-art error rates for the challenging Mask attacks (i.e., D-EER of 9.33), even though some of prior techniques~\cite{Liu-PAD-AuxSupervision-CVPR-2018,Yu-PAD-CDC-CVPR-2020} employ additional information such as depth and temporal cues to detect those 3D Mask attacks. In addition, it should be noted that the FV algorithm reports a detection performance deterioration for the BSIF filter setting adopted from the known attack evaluation (i.e., $N$ = 10 filters of size $l$ = 9), thereby resulting in a mean D-EER of 15.86\%. Despite of the accuracy degradation, this representation is still able to achieve state-of-the-art schemes, thereby showing its soundness for this scenario. In order to enhance the BSIF computation and remove the dependency to the current 60 filter configurations, we plan as future work to perform the BSIF quantisation over the filters learned by intermediate CNN layers.          

Taking a closer look at Tab.~\ref{tab:unknowPAI_SIW}, we can also see that all PAD techniques report a poor detection performance for obfuscation attacks: D-EERs in the range of 22\% - 72\% point them out as the most challenging PAI species. This is due to the fact that the makeup applied over the faces are subtle and hence look like real human faces. Given that the majority of subjects in the obfuscation set are not in the BP dataset, a proper evaluation reporting the impact of those beautifications on a real face recognition system cannot be carried out. The main question to address the threat of a given PAI is whether it is able to change the appearance of the subject enough to lead to a False-Non-Match. However, other studies about the impact of similar obfuscated images on real deep face recognition systems have reported a high biometric performance: a reliable Genuine Acceptance Rate (GAR) of 92.20\% at a False Match Rate (FMR) of 0.1\% for ArcFace~\cite{Deng-FR-Disguised-CVPR-2019} and a remarkable GAR of 98.40\% at a FMR = 0.01\% for a new ArcFace variant~\cite{Singh-FR-DisguisedFaceInWild-CVPR-2019} indicate the low dangerousness of those PAI species for facial biometric systems. Based on these observations, we think that those attacks should not be taken into account for PAD training since they can negatively impact the detection of another PAI species (e.g., Transparent Masks, see Tab.~\ref{tab:unknowPAI_SIW}). In other words, we think that by excluding obfuscation attacks from the training set, we could significantly improve the detection performance of current PAD techniques.

Finally, we observe in Fig.~\ref{fig:challenging_unk_PAIs_det} that our approach reports an average BPCER of 21.53\% for the challenging mask attacks: only one in 100 attack presentation attempts are accepted while at most 22 in 100 bona fide presentations are rejected by our PAD system. In addition, it should be noted that the proposed method achieves a remarkable BPCER of 0.0\% for any APCER over impersonation attacks, which, unlike obfuscation attacks, have reported a biometric performance deterioration for real deep face recognition systems (i.e., GAR = 52.20\% @ FMR = 0.01\%~\cite{Singh-FR-DisguisedFaceInWild-CVPR-2019}).         

\begin{table*}[!t]
	\centering
	\processtable{Benchmark with the state-of-the-art for \textit{challenging unknown PAI species} on RGB for $K$ = 1024 in terms of D-EER (\%). The best results per PAI species are highlighted in bold. \label{tab:unknowPAI_SIW}}
	{\vspace*{0.1cm}
	\begin{adjustbox}{max width=\textwidth}
	\setlength\tabcolsep{3.0pt}
	\begin{threeparttable}
	\begin{tabular}{r| c| c| c c c c c| c c c| c c c| c }  
	\toprule
	\multirow{2}{*}{Methods}		 											& \multirow{2}{*}{Replay}			&  \multirow{2}{*}{Printed}	& 							\multicolumn{5}{c|}{Mask Attacks}										&		\multicolumn{3}{c|}{Makeup Attacks}					&		\multicolumn{3}{c|}{Partial Attacks}				&  \multirow{2}{*}{Average}	\\
																				&									&							&	Half			&	Silicone		&	Trans.			&	Papercraft		&	Manneq.			&	Obfusc.			&	Imperson.		&	Cosmetic		&	Funny Eye		&	Paper Glasses	&	Partial Paper	&							\\
	\midrule
	\multirow{1}{*}{Auxiliary~\cite{Liu-PAD-AuxSupervision-CVPR-2018}}			&	14.00							&			4.30			&	11.60			&	12.40			&	24.60			&	7.80			&	10.00			&	72.10			&		10.00		&	\textbf{9.40}	&	 21.40			&		18.60		&		4.00		&		16.95 $\pm$ 17.72	\\
	
	\multirow{1}{*}{DTN~\cite{Liu-PAD-DeepTree-CVPR-2019}}						&	10.00							&		\textbf{2.10}		&	14.40			&	18.60			&	26.50			&	5.70			&	9.60			&	50.20			&		10.10		&	13.20			&	 19.80			&		20.50		&		8.80		&		16.12 $\pm$ 12.23	\\
	
	\multirow{1}{*}{DeepPixBis~\cite{George-PAD-DeepPixBis-2019}}				&	11.68							&			7.94			&	7.22			&	15.04			&	21.30			&	3.78			&	4.52			&	26.49			&		1.23		&	14.89			&	 23.28			&		18.90		&		4.82		&		12.39 $\pm$ 8.25	\\

	\multirow{1}{*}{MCCNN~\cite{George-PAD-OneClassRep-TIFS0-2020}}				&	12.82							&			12.94			&	11.33			&	13.70			&\textbf{13.47}		&	0.56			&	5.60			&	\textbf{22.17}	&		0.59		&	15.14			&\textbf{14.40}		&		23.93		&		9.82		&		12.04 $\pm$ 6.92	\\

	\multirow{1}{*}{CDCN++~\cite{Yu-PAD-CDC-CVPR-2020}}							&	\textbf{9.20}					&			5.60			&  \textbf{4.20}	&\textbf{11.10}		&	19.30			&	5.90			&	5.00			&	43.50			&	\textbf{0.00}	&	14.00			&	 23.30			&\textbf{14.30}		&	\textbf{0.00}	&		11.95 $\pm$ 11.79	\\
	\midrule
	\multirow{1}{*}{Proposed Method (Optimum)}									&	10.28							&			7.70			&	7.98			&	18.42			&	17.87			&	\textbf{0.00}	&	\textbf{2.40}	&	27.93			&	\textbf{0.00}	&	16.78			&	 17.84			&		18.22		&		3.27		&		\textbf{11.44 $\pm$ 8.73}	\\ 
	\multirow{2}{*}{Optimum BSIF filters}										&	$N$ = 11						&			$N$ = 5			&	$N$ = 7			&	$N$ = 8			&	$N$ = 10		&	$N$ = 5			&	$N$ = 6			&	$N$ = 6			&		$N$ = 6		&	$N$ = 9			&	$N$ = 9			&		$N$ = 11	&	$N$ = 11		&										\\
																				&	$l$ = 7							&			$l$ = 3			&	$l$ = 15		&	$l$ = 5			&	$l$ = 7			&	$l$ = 11		&	$l$ = 11		&	$l$ = 11		&		$l$ = 13	&	$l$ = 13		&	$l$ = 13		&		$l$ = 7		&	$l$ = 13		&										\\
	\midrule
	\multirow{1}{*}{Proposed Method (Fixed)\tnote{$\star$}}						&	12.49							&			11.76			&	14.20			&	22.94			&	23.20			&	5.61			&	7.19			&	34.57			&		1.58		&	22.07			&	 23.71			&		23.26		&		3.65		&		15.86 $\pm$ 9.89	\\
	\bottomrule
	\end{tabular}
	\begin{tablenotes}\footnotesize
		\item[$\star$] D-EERs per dataset are reported for $N$ = 10 filters of size $l$ = 9.
	\end{tablenotes}
\end{threeparttable}
\end{adjustbox}}{}
\end{table*}

\subsection{Cross-database}
\label{sec:cross_database}

Given the rapid evolution experienced by the technology, information, and communication industry, it is likely that some capture devices will be replaced by new sensors for which we have no AP samples for training the PAD systems. Therefore, it is of utmost importance that our PAD methods are robust to those situations. To that end, we select three databases (i.e., CASIA Face Anti-Spoofing, MSU-MFSD, and REPLAY-ATTACK) and establish in Tab.~\ref{tab:cross_dataset} a benchmark of our proposed representation with the current state-of-the-art techniques for each training-test configuration. It should be noted that our approach is able to achieve current state-of-the-art results, thereby yielding a D-EER of 18.24\% on average for the best BSIF filter configuration. In addition, the best performing deep learning-based scheme for cross-database (i.e., DR-UDA~\cite{Wang-PAD-AdvDomainAdaptation-TIFS-2020}) reports, on average, a D-EER of 17.93\%, which is up to twice lower than the worst result reported for this scenario (i.e., 36.83\% for DupGAN~\cite{Hu-PAD-DuplexGANs-CVPR-2018}). In order to improve generalisation cross-dataset, this method, like DupGAN~\cite{Hu-PAD-DuplexGANs-CVPR-2018}, KSA~\cite{Li-PAD-MMD-DA-TIFS-2018}, and ADA~\cite{Wang-CrossDBFacePAD-ICB-2019}, is fully based on domain adaptation, which transfers the knowledge learned from a source domain to a target domain. In spite of results attained for this scenario, the DR-UDA algorithm is unable to achieve reliable error rates for known attacks (i.e., D-EERs of 3.20\%, 6.00\%, and 7.20\% for CASIA, MSU-MFSD, and Rose-Youtu databases, respectively).

Consequently with the results reported in Tab.~\ref{tab:unknowPAI_SIW} for a fixed BSIF filter configuration (i.e., $N$ = 10 filters of size $l$ = 9) our proposed method decreases its detection performance up to 40\%, thereby resulting in a D-EER of 29.97\%. This, in turn, states the need for removing the dependency to current BSIF filters in order to keep stable the performance of our algorithm for different PAD scenarios.      

On the other hand, it may be observed that our proposed method trained with images of varying resolutions in CASIA performs well for high-resolution face images (i.e., 12.86\% for MSU). In contrast, it reports a detection performance decrease up to 47\% for face samples stemming from low-resolution capture devices (i.e., 24.36\% for REPLAY-ATTACK). However, by training our approach with high-resolution images in MSU-MFSD, a D-EER of 6.57\% can be yielded for those low-quality face images in REPLAY-ATTACK, thereby indicating the need for future studies about the impact of external factors such as image resolution and acquisition conditions over this challenging scenario. Furthermore, unlike current PAD techniques in the literature, a reliable D-EER of 11.67\% for high-resolution face images can be attained by tuning our proposed PAD algorithm with low-quality images in REPLAY-ATTACK. These results confirm that PAIs in MSU and REPLAY-ATTACK contain similar artefacts which can be successfully represented by the semantic sub-groups learned by the GMM. 

\begin{figure}[!t]
	\centering
	\includegraphics[width=0.9\linewidth]{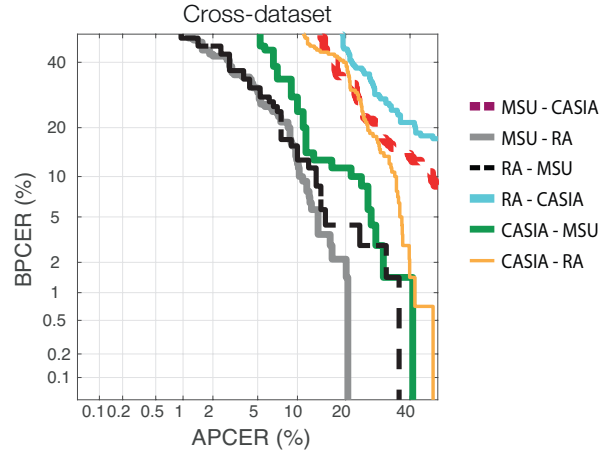}
	\caption{DET curves for the cross-dataset scenario for the best BSIF filter configurations.}
	\label{fig:cross-dataset_det}
\end{figure}

Finally, it should be noted in Fig~\ref{fig:cross-dataset_det} that the proposed algorithm reports a detection performance deterioration for high-security thresholds: a poor average BPCER100 of 73.90\% confirms the need for new interoperable PAD schemes in order to improve their generalisation capabilities for this scenario without losing accuracy for the remaining scenarios.



\begin{table}[!t]
	\centering
	\processtable{Benchmark with the state-of-the-art in terms of the D-EER (\%) for the \textit{Cross-dataset} scenarios over the best BSIF filter configuration. The best results per PAI species are highlighted in bold.\label{tab:cross_dataset}} 
	{
		\begin{adjustbox}{max width=\linewidth}
			\begin{threeparttable}
		\begin{tabular}{r|cc|cc|cc|c} \toprule
												Train				 	& 				\multicolumn{2}{c|}{MSU} 	 	 & 		\multicolumn{2}{c|}{RA} 				& 		\multicolumn{2}{c|}{CASIA}				&	\multirow{2}{*}{Avg.}	\\ 
												Test		 		 	& 		 CASIA 			&         RA    	 &    	     MSU 	 		&     	CASIA    		& 	  	  MSU 		&  	  RA  					&				    \\ \midrule
	Colour Texture~\cite{Boulkenafet-PAD-ColorTextureAnalysis-TIFS-2016} &  	46.00   			&		33.90  	 	 &      34.10  	   			&		37.70	 		& 		 24.40	  	& 	 30.30					&	34.40			\\
	Texture fusion~\cite{Boulkenafet-PAD-ColorGeneralisation-2018} 		&  	29.20   			&		16.20  	 	 &      21.40  	   			&		31.20	 		& 		 19.90	  	& 	 \textbf{9.90}			&	21.30			\\
	DupGAN~\cite{Hu-PAD-DuplexGANs-CVPR-2018} 							&  	27.10  				&		35.40 	 	 &      36.20	   			&		46.50	 		& 		 33.40  	& 	 42.4					&	36.83			\\
	KSA~\cite{Li-PAD-MMD-DA-TIFS-2018} 									&  	\textbf{9.10}  		&		33.30 	 	 &      34.90	   			&	\textbf{12.30}	 	& 		 15.10  	& 	 39.30					&	24.00			\\
	ADA~\cite{Wang-CrossDBFacePAD-ICB-2019} 							&  	17.70  				&		5.10 	 	 &      30.50	   			&		41.50	 		& 		 9.30  		& 	 17.50					&	20.27			\\
	DR-UDA~\cite{Wang-PAD-AdvDomainAdaptation-TIFS-2020} 				&  	16.80   			&		\textbf{3.00}&      29.00  	   			&		34.20	 		& 	\textbf{9.00}	& 	 15.60					&	\textbf{17.93}			\\ \midrule
					 Proposed Method (Optimum) 				   	 		&  	24.67   			&		6.57  	 	 &      \textbf{11.67}  	&		29.33	 		& 		 12.86	  	& 	 24.36					&	18.24			\\
					 \multirow{2}{*}{Optimum BSIF filters}   	 		&  	$N$	= 7  			&		$N$	= 7  	 &      $N$	= 12  	   		&		$N$	= 10	 	& 		$N$	= 8	  	& 	 $N$ = 6				&	\multirow{2}{*}{-}	\\
																		&  	$l$ = 7  			&		$l$	= 3  	 &      $l$	= 11  	   		&		$l$	= 9	 		& 		$l$	= 13	& 	 $l$ = 3				&						\\
					\midrule
					Proposed Method (Fixed)\tnote{$\star$} 				&  	31.56   			&		25.79  	 	 &      24.29  				&		29.33	 		& 		 33.10	  	& 	 35.71					&	29.97			\\ 	
																			  
					 \bottomrule
		\end{tabular}
		\begin{tablenotes}\footnotesize
			\item[$\star$] D-EERs per dataset are reported for $N$ = 10 filters of size $l$ = 9.
		\end{tablenotes}
	\end{threeparttable}
	\end{adjustbox}}{}
\end{table}

\subsection{Computational efficiency}

In the last experiments, the computational efficiency of our proposed method for $N$ = 10 filters of size $l$ = 9 and $K$ = 1024 is evaluated. Given that the computation cost of our approach depends on the number of points over the regular grid and hence the face size, we selected images with different face size ranges: small, 150~-~199~$\times$~150~-~199; medium, 250~-~350~$\times$~350~-~450; large, 550~-~700~$\times$~600~-~1050 pixels. Afterwards, the average classification time on an Intel Core i7-8750H @ 2.2 GHz, 16GB RAM was computed. As a result of this experiment, the algorithm reports an approximate average time of 0.650, 2.7, and 13.6 seconds to analyse small, medium, and large probe face samples. The efficiency value attained for large images indicates the need for enhancing the feature extraction of our proposal by decreasing the number of fixed points on the regular grid or the image size without losing accuracy. In future work, we plan to investigate different face parts such as the eye's area, mouth, and noise in order to improve both the detection performance and efficiency of our approach.  

\subsection{Failure cases}

\renewcommand{\thefootnote}{\arabic{footnote}}

Finally, we determine those failure cases which were wrongly classified by our proposed method. We noted that most misclassified BPs in SiW-M database~\footnote[1]{Samples in the SiW-M database cannot be displayed due to the database owner's requirements.} are due to the presence of heterogeneous illumination. Those BPs were classified as printed or video replay attacks, which exhibit similar non-regular bright lighting patterns. Additional artefacts such as glasses could be also affected the detection performance of our algorithm. Furthermore, most wrongly-classified APs are due to subtle patterns in the creation of these attacks. We observe that PAI species wrongly classified as bona fide presentations include Makeup or Paper glasses. Since the BSIF feature extraction is performed over the whole face image in our pipeline, we think that a suitable selection of points (e.g., landmarks) for this purpose could improve the detection of those PAI species. A local analysis of different face regions could also enhance the detection performance of our scheme.   

\section{Conclusion}
\label{sec:conclusions}

In this work, a new face PAD approach to generalise to challenging scenarios such as unknown PAI species and cross-database scenarios was proposed. In essence, this technique projects compact dense-BSIF descriptors into a new feature space, which allows discovering semantic feature sub-groups from known samples in order to improve the PAD generalisation capabilities. For computing compact dense-BSIF histograms, we adopted the strategy presented in our preliminary study~\cite{Gonzalez-PAD-FVencForFacePAD-BIOSIG-2020}. In more details, a reduction down to 95\% in the BSIF feature vector length can be achieved with no significant impact on the recognition accuracy but strongly reducing the time required for PAD analysis. 

The experimental evaluation over five freely available databases confirmed the soundness of our proposal for detecting both known and unknown PAIs. Specifically, experimental results over our pipeline indicated the statistical advantage of RGB with respect to other colour spaces for datasets having images of varying resolutions, thereby resulting in a minimum average D-EER of 0.45\% for known attack detection. In addition, a mean D-EER of 11.44\% showed the proposed PAD method soundness in the detection of unknown PAI species. In particular, the algorithm was able to yield an APCER of 26.09\% for obfuscation attacks, which is up to four times better than the ones reported by current state-of-the-art PAD techniques. Consequently, BPCER100 in the range of 0.0\% to 17\% for traditional unknown PAI species confirmed that our PAD approach is able to yield a secure and convenient system under that challenging scenario. In order to tackle generalisation issues reported for the cross-database scenario and to build a discriminative common FV feature space, we plan, as future work, to evaluate deep generative models, which have shown to be more powerful to learn data distribution than GMM.      

\section{Acknowledgements}

This research work has been funded by the DFG-ANR RESPECT Project (406880674), and the German Federal Ministry of Education and Research and the Hessian Ministry of Higher Education, Research, Science and the Arts within their joint support of the National Research Center for Applied Cybersecurity ATHENE.


\bibliographystyle{IEEEtran}
\bibliography{2021_IETBiometrics_UnkownAttacksFace_arXiv}

\end{document}